# MS²CAST: A Video Anomaly Detection Framework via Multi-Scale Spatial Cross-Attention and Efficient Spatiotemporal Networks

**Ali Safaei[a], Mohammad Kazemi[b]**

[a] Department of Computer Science, University of Milan, Milan, Italy

[b] Department of Electrical Engineering, University of Isfahan, Isfahan, Iran

Corresponding author: Mohammad Kazemi

Email address: m.kazemi@eng.ui.ac.ir

## Abstract

Video anomaly detection (VAD) is a challenging task for surveillance systems due to the scarcity and diversity of anomalous events. While semi-supervised VAD approaches have advanced the field, many rely on computationally intensive optical flow networks to capture motion dynamics. Convolutional neural networks (CNNs) are also widely used for spatial encoding, but their effective receptive field is highly local. To address these limitations, we introduce MS²CAST, an end-to-end semi-supervised predictive VAD framework. For spatial encoding, we employ a multi-scale cross-attention encoder that fuses the hierarchical feature representations from a vision transformer (MViTv2-S) and a CNN (ConvNeXt-T) to exploit their distinct receptive fields. To model spatiotemporal dynamics without optical flow estimation, we integrate two interchangeable spatiotemporal modules in the latent space. One module (SwinLSTM) incorporates a Swin Transformer, while the other (VMRNN) integrates Vision Mamba blocks. Our objective function combines L1 loss, Gradient Difference Loss (GDL), and Multi-Scale Structural Similarity Index Measure (MS-SSIM) to ensure pixel-wise accuracy, sharpness, and structural integrity. Evaluations across the UCSD Ped2, CUHK Avenue, and ShanghaiTech datasets demonstrate frame-level Receiver Operating Characteristic (ROC) Area Under the Curve (AUC) scores that are competitive with top-tier techniques while maintaining a peak GPU memory usage of under 1 GB.

**Keywords:** Video anomaly detection; Multi-scale cross-attention; VMamba; Vision transformers; MS-SSIM

## 1. Introduction

Anomaly detection in videos aims to detect events that are non-conforming to what is normal and expected across both spatial and temporal dimensions. To ensure safe environments, video anomaly detection (VAD) plays an integral role in modern surveillance systems. However, VAD is a challenging task due to the scarcity and diversity of anomalies. Beyond traditional approaches [1], [2] recent advances in deep learning have revolutionized VAD by defining different learning tasks that are based on various supervision signals. Notably, semi-supervised VAD holds a prominent position in this area of research because it bypasses the need for exhaustively collecting rare anomalous samples. Semi-supervised VAD operates on the assumption that only normal samples are available during training, so the model will only learn normal patterns [3]. Consequently, during model inference, anomalies are detected by flagging any sample that deviates from the learned patterns. Frame reconstruction, frame prediction, and a combination of both prediction and reconstruction are three commonly used paradigms for semi-supervised VAD. For the paradigm based on frame reconstruction, the model learns how to reconstruct current normal input frames during training, which can lead to a high reconstruction error for anomalous frames during inference. However, it is possible for a high-capacity model to be occasionally good at reconstructing abnormal frames with low error, thereby increasing the false negative rate. To prevent the reconstruction of anomalous frames, subsequent studies integrated a memory module into autoencoder architecture [4], [5]. However, memory modules can undesirably increase the reconstruction errors for normal frames [6]. Moreover, a memory-augmented architecture can potentially have poor performance in cross-scene applications because memory items are set up for a specific scene [6]. The frame prediction approach can effectively mitigate the overgeneralization issue associated with reconstruction-based approaches [3].

Although predictive approaches for VAD can address the limitations of reconstruction-based methods, they introduce several challenges. Due to the success of generative adversarial networks (GANs) [7] for future frame prediction [8], they are also frequently used in the VAD literature [9], [10], [11], [12], [13], [14]. However, adversarial training is often unstable and sensitive to hyperparameter choices [15]. Incorporating optical flow to enforce motion constraints is effective and widely used [6], [9], [11], [12], [16], but it is computationally intensive. Moreover, because pre-trained optical flow networks like FlowNet 2.0 [17] and RAFT [18] are usually tuned for specific types of data, using them to estimate optical flow can pose limitations on the adaptability of a VAD model depending on the dataset [19]. Lastly, while L2 loss in pixel space is widely used as one of the training objectives [9], [10], [16], [19], it can result in blurry outputs [8], [20].

Architectures based on convolutional neural networks (CNNs) are commonly used for VAD to encode both spatial features and spatiotemporal dynamics [3]. Although recent VAD studies have introduced transformer-based architectures [19], [21], approaches combining both transformers and CNNs remain underexplored. CNNs and vision transformers (ViTs) generate remarkably different feature maps [22]. While CNNs rely on local receptive fields at early layers that grow gradually, ViTs possess a significantly more global receptive field overall [22]. Recent research on feature fusion has shown promising applications of cross-attention mechanisms across different feature representations [23], [24]. This type of fusion has also been effectively used for VAD architectures in various settings [16], [19]. Cross-attention fusion mechanisms provide a way to compute the correlation between two distinct feature representations. Because anomalies typically occur at different spatial scales, a cross-attention mechanism that fuses distinct feature maps from a CNN and a ViT can enable multi-scale anomaly detection.

Beyond spatial feature extraction, predictive VAD models should also effectively capture temporal dependencies. In the broader context of spatiotemporal forecasting, SwinLSTM [25] and VMRNN [26] were proposed to improve predictive performance over ConvLSTM [27]. SwinLSTM incorporates Swin Transformer blocks [28], utilizing window-based and shifted window-based multi-head self-attention. This design achieves the representational power of global self-attention with significantly reduced computational demands. Alternatively, VMRNN integrates Vision Mamba blocks [29], employing the 2D Selective Scan (SS2D) mechanism to extend 1D sequential scanning to 2D spatial scanning, which makes state space models (SSMs) capable of processing spatial information. As alternatives to ConvLSTM, both SwinLSTM and VMRNN are computationally much more efficient and generate higher-quality frames in future frame prediction tasks [26].

In this paper, we propose $MS^2CAST$, a predictive VAD framework that integrates a spatial cross-attention mechanism to leverage the feature representations of both ViTs and CNNs for spatial encoding. Our architecture employs parallel pathways to extract multi-scale spatial feature maps from RGB frames. Then, these maps are fused via a spatial cross-attention mechanism and passed through the latent space and skip connections. To effectively capture spatiotemporal dynamics, we integrate VMRNN and SwinLSTM as interchangeable modules in the latent space. Inspired by previous research in future frame prediction [8] and image restoration [20], our loss function combines L1 loss, Multi-Scale Structural Similarity Index Measure (MS-SSIM), and Gradient Difference Loss (GDL). In summary, our main contributions are:

- We introduce $MS^2CAST$, a novel end-to-end framework for semi-supervised VAD, leveraging the distinct receptive fields of ViTs and CNNs through a hierarchical spatial cross-attention mechanism.

- We seamlessly integrate VMRNN and SwinLSTM as alternative components into our architecture to efficiently capture motion dynamics without relying on computationally intensive optical flow networks.

- We design a loss function by merging L1 loss to preserve pixel-space fidelity, MS-SSIM to maintain perceptual integrity, and GDL to generate sharp edges within predicted frames.

- Our comprehensive evaluations across three public datasets demonstrate that our method achieves strong frame-level Receiver Operating Characteristic (ROC) Area Under the Curve (AUC) scores on par with top-tier techniques while maintaining peak GPU memory usage under 1 GB.

## 2. Related works

Semi-supervised VAD techniques have evolved to capture different kinds of anomalies in various scenarios. Whether a framework relies on single-modal or multi-modal input affects both its detection accuracy and computational demands. This section reviews the literature across these two input settings to contextualize the design of our proposed framework.

## 2.1. Multi-modal approaches

To capture complex motion dynamics, a prominent category of VAD frameworks relies on multi-modal inputs, particularly pairing optical flow with RGB frames. In general, multi-modal VAD frameworks utilize either traditional optical flow algorithms [30], [31], [32] or deep pre-trained optical flow networks [16], [33], [34]. While traditional algorithms are lightweight, they often lack the accuracy of modern deep pre-trained optical flow networks. However, the performance of these deep models can degrade when applied to domains that differ from their training data [19].

Among traditional optical flow estimation approaches, the TV-L1 algorithm [35] is widely used. For instance, Wu et al. [30] introduced an encoder-decoder architecture to process modalities independently via 3D CNNs. Their framework linearly fuses these features in the latent space to maintain a compact latent representation while reconstructing both motion and appearance. Utilizing the same TV-L1 algorithm [35], Rai et al. [31] employed two isolated 3D CNN-based autoencoders alongside pseudo-anomaly generation for the separate reconstruction of optical flow and RGB frames. Furthermore, they leveraged features from a pre-trained ViFi-CLIP [36] model to train a discriminator, providing a semantic anomaly detection criterion during inference. Focusing on computational speed, Wu et al. [32] proposed a sparse coding framework that estimates optical flow based on the method introduced in [37]. Using an adapted BN-Inception network [38], their model processes both modalities independently and fuses the spatial and temporal feature maps via channel-wise concatenation and a convolutional operation. Ultimately, it detects anomalies based on deviations from a learned normal dictionary.

To overcome the accuracy limitations of traditional algorithms, several frameworks incorporate deep pre-trained optical flow networks. AMMC-Net [34] maintains semantic consistency between appearance and motion features by utilizing feed-forward neural networks to integrate a cross-modality mapping mechanism within the latent space. This mechanism maps appearance features to motion features and vice versa. Furthermore, the network integrates memory modules in the latent space to detect anomalies based on both prediction errors and discrepancies from memory prototypes. Employing a similar anomaly detection criterion, Ning et al. [33] proposed a predictive encoder-decoder framework utilizing the pre-trained FlowNet 2.0 [17] model alongside a memory module in the latent space. They incorporated a channel attention mechanism into the motion encoder and refined the appearance features via element-wise multiplication with an attention map obtained from the fusion of spatial and temporal feature maps. Taking a different predictive approach, Ma et al. [16] leveraged a pre-trained RAFT [18] model with a cross-attention fusion mechanism for the motion-guided representation of appearance features, followed by a memory module and a decoder equipped with channel and spatial attention mechanisms. It is noteworthy that memory modules, despite being commonly integrated into VAD architectures [16], [33], [34], can limit cross-scene generalization [6]. To bypass the computational demands of optical flow estimation, many frameworks adopt single-modal architectures.

### 2.2. Single-modal approaches

Relying exclusively on RGB information to process both spatial and temporal representations requires careful architectural design to establish a cohesive single-modal VAD framework. A foundational category of these single-modal strategies relies on architectures that employ autoencoders for frame reconstruction [4], [39], [40]. For instance, Luo et al. [40] integrated ConvLSTM [27] into a convolutional autoencoder to address the motion-modeling limitations of ConvAE [39]. To prevent the reconstruction of anomalous frames, Gong et al. [4] introduced a memory-augmented architecture in which the encoded input retrieves a sparse set of stored normal patterns for reconstruction, ensuring that anomalous events yield a high reconstruction error. To capture more diverse patterns than single-encoder convolutional autoencoders, MESDnet [41] introduces an end-to-end multi-encoder architecture. It utilizes one encoder to capture motion via the difference between consecutive frames, alongside another encoder that extracts spatial information. The resulting feature maps are concatenated along the channel dimension and passed to the decoder to reconstruct the current frame. While MESDnet processes only RGB frames as inputs, it notably incorporates an adversarial loss and an optical flow loss into its training objective.

Predictive single-modal frameworks widely employ a U-Net alongside an adversarial training strategy and optical flow-based motion constraints [9], [11], [12]. Adopting an adversarial training strategy, Liu et al. [9] proposed a foundational VAD framework that employs a U-Net for future frame prediction while leveraging a per-trained optical flow network to impose motion constraints during training. Building on Frame-Pred [9], Luo et al. [11] adapted a U-Net by using a graph neural network in the latent space and the Mask R-CNN [42] object detector to model foreground visual relationships. However, object detectors can struggle to detect small objects and may underperform in crowded scenes with occluded objects, whereas choosing an object detector with higher detection accuracy usually comes at the cost of speed and efficiency [43]. To prioritize foreground elements without object-level detection, another architecture [6] integrated a spatial attention mechanism into a U-Net skip connections alongside a channel attention mechanism for spatial encoding. Exploiting rich semantic features of a latent space, Zhang et al. [12] adopted a predictive strategy across multiple spaces by not only predicting the future frame in the pixel space but also encoding those predictions into a latent space to compare them against the corresponding ground truth. Inspired by research in neuroscience, Ye et al. [44] proposed a dual-module architecture consisting of a predictive coding module that uses ConvLSTM [27] and a U-Net to encode future frames, coupled with a refinement module to model the prediction error. Following a hybrid approach while adopting an adversarial training strategy, Tang et al. [10] combined frame reconstruction and prediction strategies using only RGB frames, comparing their work against Frame-Pred [9].

Over the past few years, architectures utilizing transformers have emerged within single-modal VAD. As one of the early VAD frameworks to leverage transformers, TransAnomaly [21] adopts an adversarial training strategy and incorporates a video vision transformer into the latent space of a U-Net to serve as a spatiotemporal module capturing motion dynamics. Additionally, Tran et al. [19] proposed a predictive VAD framework utilizing transformers for aerial videos. They designed a spatial transformer encoder to extract spatial information and a temporal transformer encoder to model temporal relationships across consecutive frames, introducing a temporal cross-attention mechanism to fuse the obtained temporal tokens into a single temporal token.

While these single-modal approaches have advanced the field, many still rely on adversarial training, object detectors, or an optical flow loss. To employ an effective streamlined optimization process, we instead adopt an objective function that combines MS-SSIM, GDL, and L1 loss. Furthermore, rather than relying solely on CNNs or transformers, we leverage a spatial cross-attention mechanism to fuse the distinct feature maps of both a ViT and a CNN and pair this spatial encoding with an efficient spatiotemporal module to capture motion dynamics.

# 3. Proposed method

## 3.1. Overall architecture

We propose MS$^2$CAST, a semi-supervised framework for VAD that leverages a ViT and a CNN to extract spatial features and fuses them using a cross-attention mechanism. As illustrated in Fig. 1, the architecture includes three main components: a multi-scale cross-attention encoder, a spatiotemporal module, and a decoder. It takes a video clip consisting of four frames ($I_{t-3:t}$) to predict the next frame ($\hat{I}_{t+1}$). Grounded in the U-Net architecture [45], it first extracts two distinct sets of multi-level spatial features from each input frame using the ViT and CNN independently. It then fuses them at each level, using the CNN features to guide the feature maps obtained from the ViT. The fused feature map at the lowest resolution passes through the spatiotemporal module and then into the decoder, while the remaining fused feature maps from the last input frame ($I_t$) are used in skip connections to pass spatial information at varying resolutions to the decoder for prediction.

## 3.2. Multi-scale cross-attention encoder

To effectively capture a broad range of spatial patterns in video frames, our spatial encoder employs two distinct parallel pathways. We choose MViTv2-S [46], a multi-scale vision transformer, and ConvNeXt-T [47], a modern convolutional neural network, as our spatial feature extractors. We selected these architectures to generate the multi-scale representations required for future frame prediction in our U-Net-based framework, while leveraging the distinct receptive fields of a ViT and a CNN. High-capacity feature extractors are often computationally heavy; however, the “Tiny” variant of ConvNeXt and the “Small” variant of MViT-v2 provide an optimal balance between computational requirements and representational power. Additionally, they inherently generate multi-scale feature representations with matching resolutions at each stage. We exploit these attributes in these two architectures to fuse their feature maps at different scales.

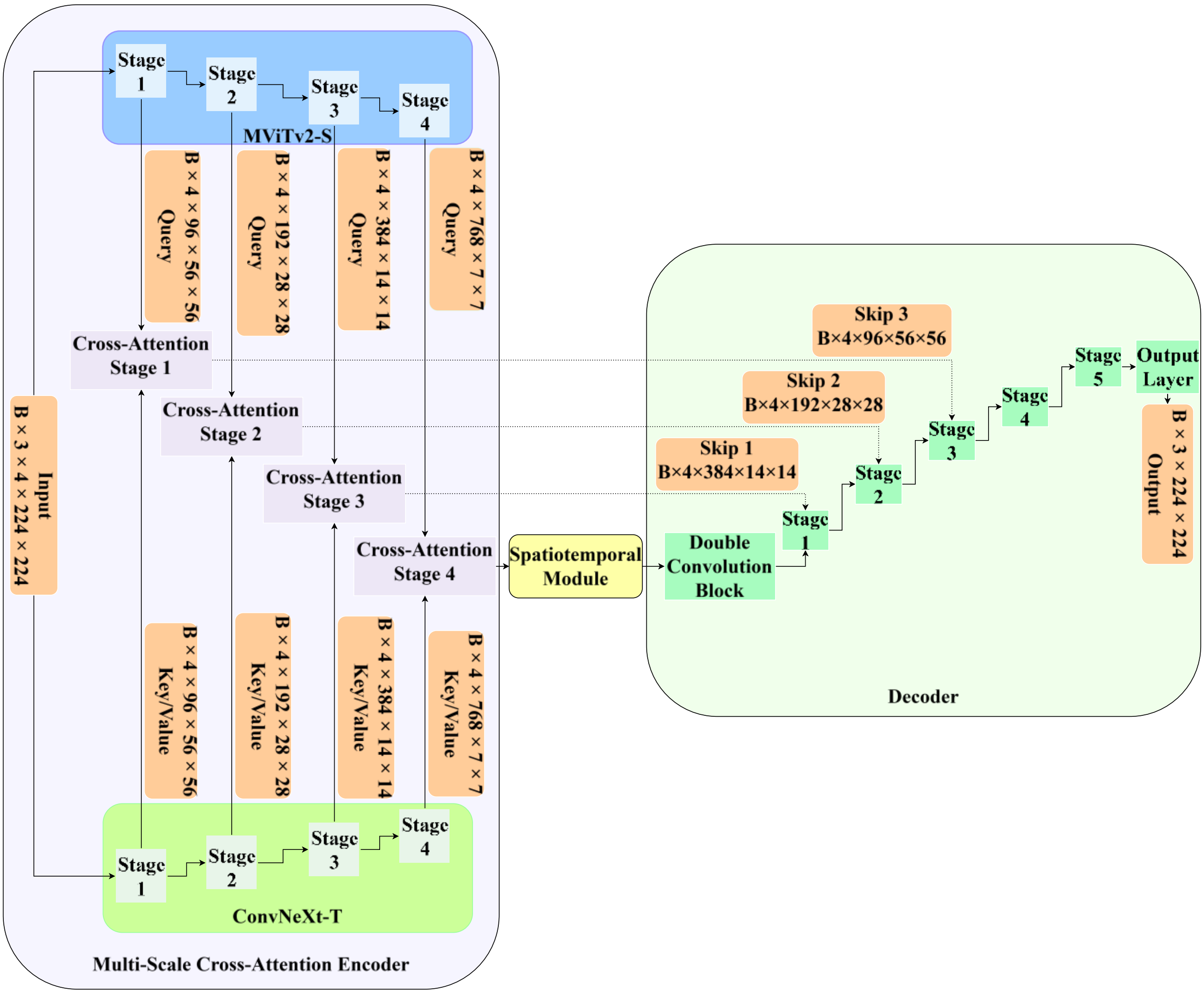

**Fig. 1.** Overview of the proposed MS²CAST framework. The architecture takes a clip of four consecutive frames to predict the future frame by leveraging a multi-scale cross-attention encoder, a spatiotemporal module, and a decoder.

### 3.2.1. Multi-scale spatial feature extractors

MViTv2 [46] is a hierarchical vision transformer that integrates decomposed relative positional embeddings and residual pooling connections alongside a hybrid window and pooling attention to address the high computational demands of global self-attention while maintaining high performance. As the other feature extractor in our spatial encoder, ConvNeXt [47] incorporates inverted bottlenecks, large kernels, and depthwise convolution. Both ConvNeXt and MViTv2 gradually reduce the spatial resolution of the input frames while expanding the channel dimension. The spatial downsampling ratios of these two networks mirror each other. At each stage, the outputs from ConvNeXt-T and MViTv2-S share perfectly aligned spatial dimensions and feature depths. We extract the intermediate feature maps from four distinct structural stages across both models and apply the cross-attention mechanism to the paired feature maps. Both backbones are initialized with ImageNet-1k [48] pre-trained weights and kept frozen during training.

Given an input sequence of $T = 4$ RGB frames $\mathbf{I} \in \mathbb{R}^{B\times T\times 3\times H\times W}$ within a batch with size of $B$, where $H = 224$ and $W = 224$, both models extract features across four hierarchical stages $l \in \{1,2,3,4\}$. The output features at stage $l$ for MViTv2-S and ConvNeXt-T are defined as $\mathbf{F}^{(l)}_{\text{MViTv2-S}} \in \mathbb{R}^{B\times T\times C_l\times H_l\times W_l}$ and $\mathbf{F}^{(l)}_{\text{ConvNeXt-T}} \in \mathbb{R}^{B\times T\times C_l\times H_l\times W_l}$, with the specific channel and spatial dimensions at each stage $l$ defined as:

- **Stage 1 ($l = 1$):** $C_1 = 96, \quad H_1 = W_1 = 56$
- **Stage 2 ($l = 2$):** $C_2 = 192, \quad H_2 = W_2 = 28$
- **Stage 3 ($l = 3$):** $C_3 = 384, \quad H_3 = W_3 = 14$
- **Stage 4 ($l = 4$):** $C_4 = 768, \quad H_4 = W_4 = 7$

### 3.2.2. 2D sinusoidal spatial positional encoding

The cross-attention fusion used in this study operates on flattened sequences of spatial feature maps. This characteristic makes it an inherently permutation-invariant mechanism. To preserve 2D spatial topology across spatial feature maps, we utilize a non-learnable 2D sinusoidal positional encoding [49], adapted from the foundational 1D sequence formulation [50]. Considering coordinates $(x, y, c)$ where $x \in \{0, \dots, W_l - 1\}$, $y \in \{0, \dots, H_l - 1\}$ and $c \in \{0, \dots, C_l - 1\}$, the channel dimension $C_l$ is partitioned equally into $x$-axis and $y$-axis representations. For a given position $(x, y, c)$ and channel indices $k \in \{0, \dots, C_l/4 - 1\}$, the frequency constant $\omega_k$ and the encodings are defined as:

$$\omega_k = \left(\frac{1}{10000}\right)^{4k/C_l} \tag{1}$$

$$\text{PE}_l(x, y, c) = \begin{cases} \sin(y \cdot \omega_k), & \text{if } c = 2k \\ \cos(y \cdot \omega_k), & \text{if } c = 2k + 1 \\ \sin(x \cdot \omega_k), & \text{if } c = C_l/2 + 2k \\ \cos(x \cdot \omega_k), & \text{if } c = C_l/2 + 2k + 1 \end{cases} \tag{2}$$

This encoding is scaled dynamically to fit the resolution $H_l \times W_l$ at each stage and broadcast across the batch and temporal dimensions.

### 3.2.3. Multi-head cross-attention mechanism

The feature maps at each stage $l$ are flattened into representations of shape $\mathbb{R}^{BT \times H_l W_l \times C_l}$ and subsequently passed into the cross-attention mechanism. Considering flattened representations $\mathbf{Z}^{(l)}_{\text{MViTv2-S}} \in \mathbb{R}^{BT \times H_l W_l \times C_l}$ and $\mathbf{Z}^{(l)}_{\text{ConvNeXt-T}} \in \mathbb{R}^{BT \times H_l W_l \times C_l}$, we define Query ($\mathbf{Q}$), Key ($\mathbf{K}$), and Value ($\mathbf{V}$) for our cross-attention mechanism at each stage $l$:

$$\mathbf{Q}^{(l)} = \mathbf{Z}^{(l)}_{\text{MViTv2-S}}, \quad \mathbf{K}^{(l)} = \mathbf{Z}^{(l)}_{\text{ConvNeXt-T}}, \quad \mathbf{V}^{(l)} = \mathbf{Z}^{(l)}_{\text{ConvNeXt-T}} \tag{3}$$

Using $h = 6$ parallel attention heads, the head projection dimension is $d_h = C_l/h$. For head $i \in \{1, \dots, h\}$, linear transformations with learnable weights $\mathbf{W}_i^Q, \mathbf{W}_i^K, \mathbf{W}_i^V \in \mathbb{R}^{C_l \times d_h}$ project the sequence representations into subspace projections:

$$\mathbf{Q}_i = \mathbf{Q}^{(l)}\mathbf{W}_i^Q, \quad \mathbf{K}_i = \mathbf{K}^{(l)}\mathbf{W}_i^K, \quad \mathbf{V}_i = \mathbf{V}^{(l)}\mathbf{W}_i^V \tag{4}$$

The Multi-Head Attention (MHA) operation across all heads is formulated as:

$$\text{Head}_i = \text{Softmax}\left(\frac{\mathbf{Q}_i \mathbf{K}_i^\top}{\sqrt{d_h}}\right)\mathbf{V}_i \tag{5}$$

$$\text{MHA}\left(\mathbf{Q}^{(l)}, \mathbf{K}^{(l)}, \mathbf{V}^{(l)}\right) = \text{Concat}(\text{Head}_1, \text{Head}_2, \dots, \text{Head}_h)\mathbf{W}^O \tag{6}$$

where $\mathbf{W}^O \in \mathbb{R}^{C_l \times C_l}$ represents the output projection matrix.

In transformer architecture, layer normalization is traditionally applied after the residual connections. This configuration is known as Post-Layer Normalization (Post-LN). In the Post-LN configuration, positional encodings are directly added to raw inputs, and layer normalization is applied after residual summation:

$$\mathbf{Q}^{(l)}_{\text{pe}} = \mathbf{Q}^{(l)} + \mathbf{PE}^{(l)}, \quad \mathbf{K}^{(l)}_{\text{pe}} = \mathbf{K}^{(l)} + \mathbf{PE}^{(l)} \tag{7}$$

$$\mathbf{A}^{(l)} = \mathrm{MHA}\left(\mathbf{Q}_{\mathrm{pe}}^{(l)}, \mathbf{K}_{\mathrm{pe}}^{(l)}, \mathbf{V}^{(l)}\right) \tag{8}$$

$$\mathbf{Z}_{\mathrm{intermediate}}^{(l)} = \mathbf{Q}^{(l)} + \mathbf{A}^{(l)} \tag{9}$$

$$\mathbf{Z}_{\mathrm{fused}}^{(l)} = \mathrm{LN}\left(\mathbf{Z}_{\mathrm{intermediate}}^{(l)}\right) \tag{10}$$

There is also another configuration called Pre-Layer Normalization (Pre-LN) in which layer normalizations $\mathrm{LN}_Q$ and $\mathrm{LN}_K$ are applied to Query and Key/Value representations prior to the positional encoding process and attention computation:

$$\mathbf{Q}_{\mathrm{norm}}^{(l)} = \mathrm{LN}_Q\left(\mathbf{Q}^{(l)}\right), \quad \mathbf{K}_{\mathrm{norm}}^{(l)} = \mathrm{LN}_K\left(\mathbf{K}^{(l)}\right), \quad \mathbf{V}_{\mathrm{norm}}^{(l)} = \mathbf{K}_{\mathrm{norm}}^{(l)} \tag{11}$$

$$\mathbf{Q}_{\mathrm{pe}}^{(l)} = \mathbf{Q}_{\mathrm{norm}}^{(l)} + \mathbf{PE}^{(l)}, \quad \mathbf{K}_{\mathrm{pe}}^{(l)} = \mathbf{K}_{\mathrm{norm}}^{(l)} + \mathbf{PE}^{(l)} \tag{12}$$

$$\mathbf{A}^{(l)} = \mathrm{MHA}\left(\mathbf{Q}_{\mathrm{pe}}^{(l)}, \mathbf{K}_{\mathrm{pe}}^{(l)}, \mathbf{V}_{\mathrm{norm}}^{(l)}\right) \tag{13}$$

$$\mathbf{Z}_{\mathrm{fused}}^{(l)} = \mathbf{Q}^{(l)} + \mathbf{A}^{(l)} \tag{14}$$

In our experiments, we tested both the Pre-LN (Fig. 2(a)) and Post-LN (Fig. 2(b)) cross-attention configurations to compare their effects on our framework. Experimental details regarding these configurations are discussed in Section 4.

Following cross-attention fusion, the flattened sequence $\mathbf{Z}_{\mathrm{fused}}^{(l)} \in \mathbb{R}^{BT \times H_l W_l \times C_l}$ is reshaped back to spatial feature dimension $\mathbf{F}_{\mathrm{fused}}^{(l)} \in \mathbb{R}^{B \times T \times C_l \times H_l \times W_l}$. The lowest-resolution feature map ($l = 4$) is forwarded to the spatiotemporal module, while the spatial feature maps obtained from the last input frame ($I_t$) at the first three stages are passed via skip connections to the decoder.

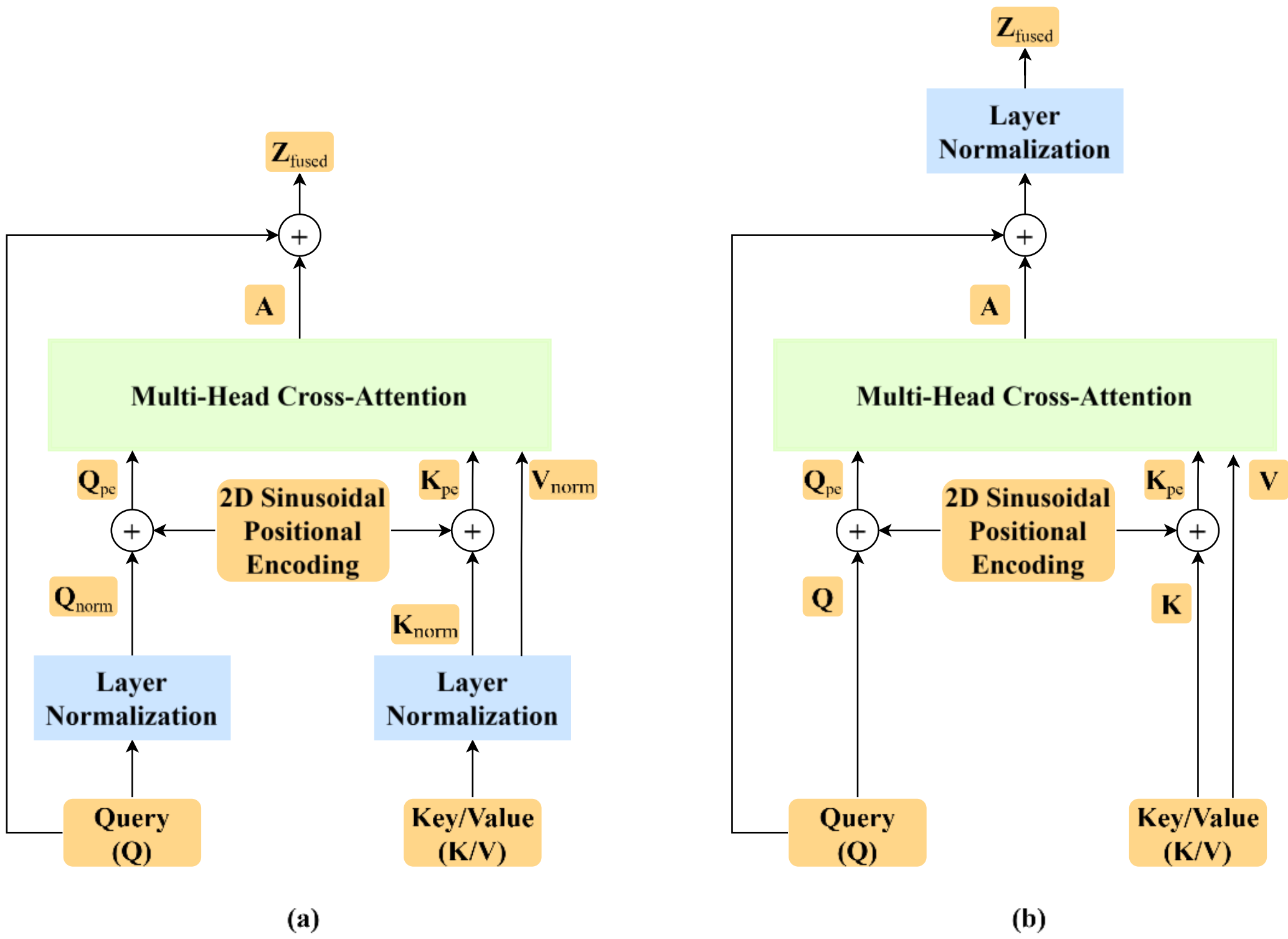

**Fig. 2.** Structural configurations for the multi-head cross-attention fusion mechanism. The diagrams illustrate the (a) Pre-Layer Normalization (Pre-LN) and (b) Post-Layer Normalization (Post-LN) configurations.

## 3.3. Spatiotemporal module

To model motion dynamics across consecutive frames, the bottleneck features $\mathbf{F}_{\text{fused}}^{(4)}$ extracted by the multi-scale cross-attention encoder, are fed into a spatiotemporal module.

The effective receptive field in deep CNNs is highly local [51]. This attribute poses a constraint on ConvLSTM [27], as its reliance on CNNs limits its ability to capture long-range spatial dependencies during spatiotemporal forecasting [26]. SwinLSTM [25] and VMRNN [26] are two alternatives to ConvLSTM that are not only significantly lighter computationally but also possess a global receptive field [26]. Furthermore, they generate higher-quality predictions compared to ConvLSTM when employed in future frame prediction tasks [26].

### 3.3.1. SwinLSTM

SwinLSTM [25] integrates Swin Transformer blocks [28] to leverage window-based and shifted window-based multi-head self-attention. At each timestep $t$, the fused spatial feature map $\mathbf{F}_{\text{fused}, t}^{(4)} \in \mathbb{R}^{B \times C_4 \times H_4 \times W_4}$ is first projected into a sequence of spatial tokens via a patch embedding layer with a $1 \times 1$ patch size:

$$\mathbf{X}_t = \text{PatchEmbed}\left(\mathbf{F}_{\text{fused}, t}^{(4)}\right) \in \mathbb{R}^{B \times H_4 W_4 \times C_4} \tag{15}$$

where $H_4 W_4 = 49$ represents the spatial token count and $C_4 = 768$ denotes the feature dimension.

To propagate temporal memory, the input token sequence $\mathbf{X}_t$ and the previous hidden state $\mathbf{H}_{t-1} \in \mathbb{R}^{B \times H_4 W_4 \times C_4}$ are fused at the entry of the first Swin Transformer block. Specifically, both states are normalized using layer normalization and concatenated along the channel dimension:

$$\mathbf{Z}_t^{(0)} = \mathbf{W}[\text{LN}(\mathbf{X}_t) \parallel \text{LN}(\mathbf{H}_{t-1})] \tag{16}$$

where $\mathbf{W} \in \mathbb{R}^{C_4 \times 2C_4}$ is a learnable projection matrix and $\parallel$ represents channel-wise concatenation.

The combined state $\mathbf{Z}_t^{(0)}$ is sequentially processed through a cascade of $D = 6$ Swin Transformer blocks, each equipped with residual connections. To continuously guide the temporal representations with current spatial information, the initial token sequence $\mathbf{X}_t$ is concatenated with the intermediate feature maps at the entry of every odd-indexed block $m > 1$, where $m \in \{1, \ldots, D\}$. The $m$-th Swin Transformer block alternates between window-based multi-head self-attention (W-MSA) when $m$ is odd and shifted window-based multi-head self-attention (SW-MSA) when $m$ is even. For a local window of size $M \times M$, where $M = 7$, the self-attention operation is defined as:

$$\text{Attention}(\mathbf{Q}, \mathbf{K}, \mathbf{V}) = \text{Softmax}\left(\frac{\mathbf{Q}\mathbf{K}^\top}{\sqrt{d}} + \mathbf{B}_{\text{rel}}\right)\mathbf{V} \tag{17}$$

where $\mathbf{Q}, \mathbf{K}, \mathbf{V} \in \mathbb{R}^{M^2 \times d}$ are Query, Key, and Value matrices projected from token representations, $d = C_4/h$ is the dimension of Query/Key across $h = 24$ attention heads, and $\mathbf{B}_{\text{rel}} \in \mathbb{R}^{M^2 \times M^2}$ is a learnable relative position bias matrix.

After passing through all D blocks, the transformed representation $\mathbf{F}_t = \mathbf{Z}_t^{(D)} \in \mathbb{R}^{B \times H_4 W_4 \times C_4}$ updates the cell state $\mathbf{C}_t$ and hidden state $\mathbf{H}_t$ through a recurrent gating mechanism:

$$\mathbf{G}_t = \sigma(\mathbf{F}_t) \tag{18}$$

$$\tilde{\mathbf{C}}_t = \tanh(\mathbf{F}_t) \tag{19}$$

$$\mathbf{C}_t = \mathbf{G}_t \odot \left(\mathbf{C}_{t-1} + \tilde{\mathbf{C}}_t\right) \tag{20}$$

$$\mathbf{H}_t = \mathbf{G}_t \odot \tanh(\mathbf{C}_t) \tag{21}$$

where $\sigma(\cdot)$ denotes the sigmoid function, $\odot$ represents elementwise Hadamard multiplication, and initial states are set to zero.

### 3.3.2. VMRNN

Considering a continuous-time state space model with an input $x(t) \in \mathbb{R}$, an output $y(t) \in \mathbb{R}$, and a hidden state vector $h(t) \in \mathbb{R}^{N_{\text{state}}}$, where $N_{\text{state}} = 16$, the state space model equations are defined as [29]:

$$h'(t) = \mathbf{A}h(t) + \mathbf{B}x(t) \tag{22}$$

$$y(t) = \mathbf{C}h(t) + \mathbf{D}x(t) \tag{23}$$

where $\mathbf{A} \in \mathbb{R}^{N_{\text{state}} \times N_{\text{state}}}$, $\mathbf{B} \in \mathbb{R}^{N_{\text{state}} \times 1}$, $\mathbf{C} \in \mathbb{R}^{1 \times N_{\text{state}}}$, and $\mathbf{D} \in \mathbb{R}^1$ represent the state space parameters. To apply this continuous formulation to discrete input sequences in a deep learning framework, $\mathbf{A}$ and $\mathbf{B}$ are discretized to $\overline{\mathbf{A}}$ and $\overline{\mathbf{B}}$, respectively, by using a zero-order hold (ZOH) approach with a timescale parameter [29]. This results in the following discrete-time equations at sequence index $k$:

$$h_k = \overline{\mathbf{A}}h_{k-1} + \overline{\mathbf{B}}x_k \tag{24}$$

$$y_k = \mathbf{C}h_k + \mathbf{D}x_k \tag{25}$$

The core architectural component of VMRNN [26] is the Visual State Space block (VSB) [29]. At timestep $t$, the fused spatial feature map $\mathbf{F}_{\text{fused}, t}^{(4)}$ is first projected into a sequence of spatial tokens. To propagate temporal memory, this sequence is concatenated with the previous hidden state $\mathbf{H}_{t-1}$ along the channel dimension and linearly projected at the entry of the first VSB, as formulated in Eqs. (15) and (16). In the VSB, the tokens are reshaped back to their 2D spatial orientation. The features are then linearly projected to an expanded inner dimension $D_{\text{inner}} = 2C_4$ and split into two parallel streams $\mathbf{X}_{\text{in}} \in \mathbb{R}^{B\times H_4\times W_4\times D_{\text{inner}}}$ and $\mathbf{Z}_{\text{in}} \in \mathbb{R}^{B\times H_4\times W_4\times D_{\text{inner}}}$. Then, the stream $\mathbf{X}_{\text{in}}$ is transposed to a tensor of shape $\mathbb{R}^{B\times D_{\text{inner}}\times H_4\times W_4}$ and passes through a $3 \times 3$ depthwise convolution followed by a SiLU activation function. Employing the 2D Selective Scan (SS2D) mechanism, the result is scanned along four distinct trajectories to generate 1D sequence streams $\mathbf{x}^{(v)} \in \mathbb{R}^{B\times D_{\text{inner}}\times H_4 W_4}$, where $v \in \{1, 2, 3, 4\}$: raster scan, reversed raster scan, column-major scan, reversed column-major scan. Each sequence $\mathbf{x}^{(v)}$ is processed by an independent discrete SSM to ensure spatial context aggregation from all directions. All the outputs $\mathbf{y}^{(v)}$ are restored to their original 2D spatial orientations via inverse scanning operators and then aggregated by summation. Finally, the aggregated output $\mathbf{Y}_{\text{SS2D}}$ is linearly projected back to dimension $C_4$:

$$\mathbf{Y}_{\text{out}} = \mathbf{W}_{\text{out}}\big(\text{LN}(\mathbf{Y}_{\text{SS2D}}) \odot \text{SiLU}(\mathbf{Z}_{\text{in}})\big) \tag{26}$$

The initial input passes through a network that integrates $D = 6$ VSBs equipped with residual connections. After passing through all $D$ blocks, the final block output $\mathbf{F}_t = \mathbf{Z}_t^{(D)} \in \mathbb{R}^{B\times H_4 W_4\times C_4}$ passes through the exact same recurrent gating mechanism described for SwinLSTM in Eqs. (18)–(21) to update the cell state $\mathbf{C}_t$ and hidden state $\mathbf{H}_t$ [26].

## 3.4. Decoder

As the final architectural component of MS$^2$CAST, the decoder is designed to generate the future frame $\hat{I}_{t+1}$ by mapping the final hidden state $\mathbf{H}_t$ produced by the spatiotemporal module. The decoder utilizes five distinct upsampling stages to generate output frames with spatial dimensions of $224 \times 224$.

Prior to upsampling, $\mathbf{H}_t$ is processed by a Double Convolution block, which is composed of two identical sequences of operations, each consisting of a $3 \times 3$ convolution followed by a group normalization layer and a ReLU activation function. As illustrated in Fig. 3, the decoder employs a hierarchical upsampling strategy to generate the output frame:

- **Stages 1 to 3 (With Skip Connections):** Each of the first three upsampling stages applies bilinear interpolation with a scale factor of 2 to expand the spatial dimensions of a feature map, immediately followed by a $3 \times 3$ convolution to reduce the channel dimension. These convolution results are concatenated along the channel dimension with the corresponding fused intermediate feature maps obtained from the final input frame $\mathbf{I}_t$ in the multi-scale cross-attention encoder, including $\mathbf{F}_{\text{fused},\,t}^{(3)} \in \mathbb{R}^{B\times C_3\times H_3\times W_3}$ for the first stage, $\mathbf{F}_{\text{fused},\,t}^{(2)} \in \mathbb{R}^{B\times C_2\times H_2\times W_2}$ for the second, and $\mathbf{F}_{\text{fused},\,t}^{(1)} \in \mathbb{R}^{B\times C_1\times H_1\times W_1}$ for the third. After concatenation, the obtained representations are processed by a Double Convolution block.

- **Stages 4 to 5 (Without Skip Connections):** The fourth and fifth upsampling stages operate without skip connections, relying entirely on all the steps introduced for the first three stages except concatenation.

- **Output Layer:** The last upsampling stage results in a feature map with 24 channels and $224 \times 224$ spatial dimensions. Subsequently, the output layer applies a $1 \times 1$ convolution to reduce the channel dimension to 3, resulting in the RGB representation. To generate the predicted frame, a sigmoid activation function is then applied to map the output values to a continuous range between 0 and 1.

We chose group normalization over batch normalization in the Double Convolution blocks because batch normalization relies on the statistical mean and variance across the batch dimension. This reliance can cause performance degradation if the batch statistics do not accurately represent the actual dataset domain statistics. Conversely, group normalization calculates the mean and variance independently for each channel group within an individual sample, which makes the normalization process inherently independent of batch size.

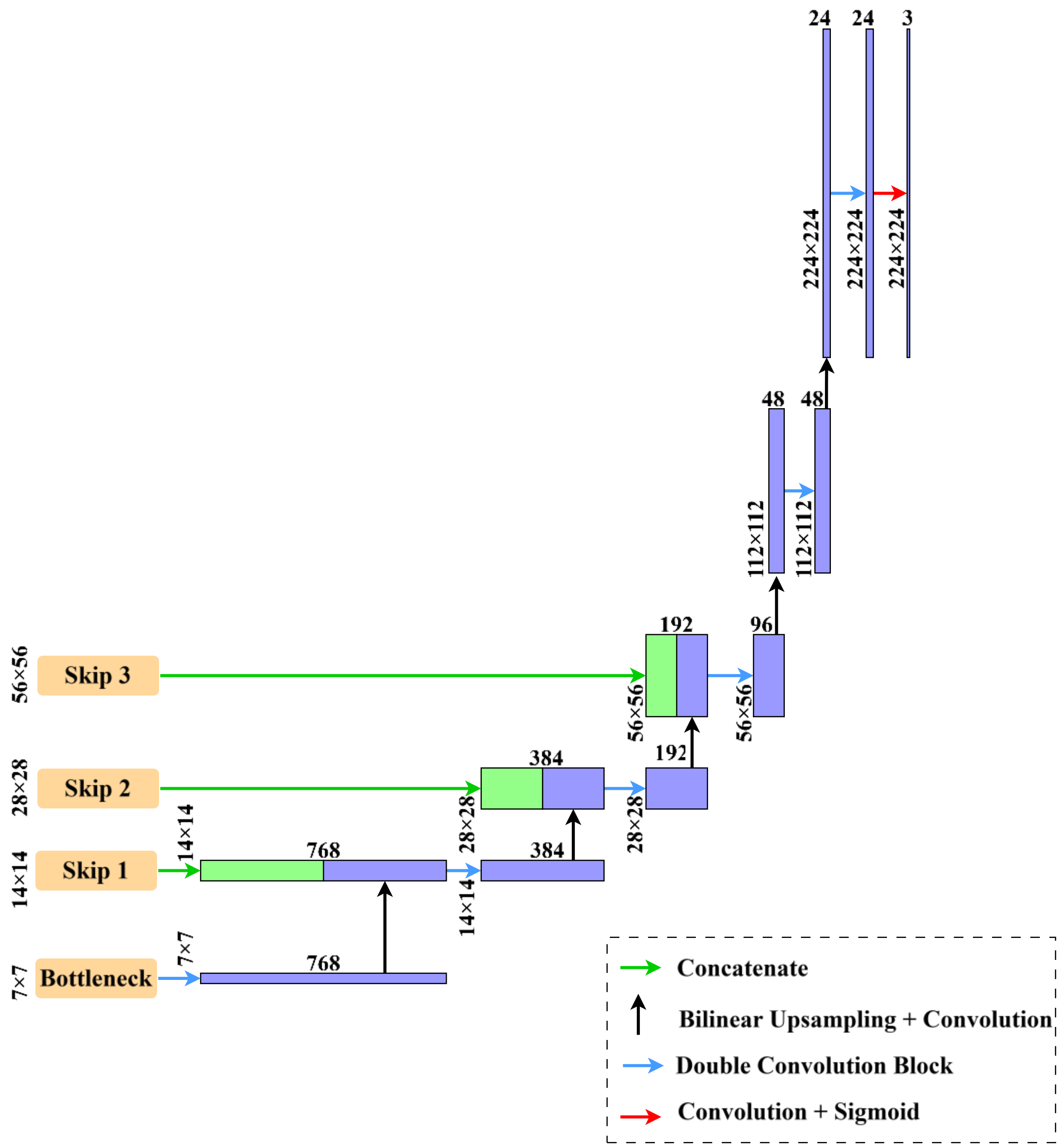

**Fig. 3.** The hierarchical upsampling strategy employed by the decoder to generate the predicted frame.

### 3.5. Training loss

To train our predictive model to generate sharp frames with high quality, we adopt an objective function that integrates different criteria focused on pixel-wise accuracy, sharpness of edges, and perceptual integrity. The selection of these specific criteria is driven by the findings of both Mathieu et al. [8] and Zhao et al. [20]. To elaborate, Mathieu et al. [8] demonstrated that choosing L1 loss over L2 loss for future frame prediction yields sharper predictions with higher quality, while pairing L1 loss with GDL is an improvement over using L1 loss alone. Similarly, Zhao et al. [20] showed that employing L1 loss instead of L2 loss is a more effective strategy for super-resolution, leading to higher-quality output images. Additionally, they highlighted that a combination of L1 loss and MS-SSIM for their total loss function yields better results than using either loss function alone. Building on these established insights, we formulate our objective function.

Given a predicted frame $\hat{I}_{t+1}$ and its corresponding ground truth $I_{t+1}$, the total loss function is a weighted combination of three distinct components including L1 loss ($L_1$), GDL ($L_{\text{GDL}}$), and MS-SSIM loss ($L_{\text{MS_SSIM}}$).

The L1 loss computes the pixel-wise mean absolute difference between the predicted frame and the ground truth frame:

$$L_1 = \frac{1}{N}\left\|\hat{I}_{t+1} - I_{t+1}\right\|_1 \tag{27}$$

where $\| . \|_1$ denotes the L1 norm and $N$ represents the total number of pixels in the frame.

To generate sharp edges within a predicted frame, GDL penalizes the mean absolute error between the absolute spatial gradients of the prediction and the target along both horizontal and vertical axes:

$$L_{GDL} = \frac{1}{N}\left(\left\||\nabla_x I_{t+1}| - |\nabla_x \hat{I}_{t+1}|\right\|_1 + \left\||\nabla_y I_{t+1}| - |\nabla_y \hat{I}_{t+1}|\right\|_1\right) \tag{28}$$

where $\| . \|$ denotes the L1 norm and $| . |$ denotes the element-wise absolute value, and $N$ is the total number of pixels. The operators $\nabla_x$ and $\nabla_y$ compute the spatial gradients of the frames along the horizontal and vertical axes, respectively.

To consider the holistic structural integrity of the generated frame, we utilize MS-SSIM. The Structural Similarity Index Measure (SSIM) evaluates frame quality using three components: luminance ($l$), contrast ($c$), and structure ($s$). Considering a local window $\mathbf{x}$ in the predicted future frame and a local window $\mathbf{y}$ in the actual future frame, both extracted via an $11 \times 11$ sliding Gaussian kernel with standard deviation $\sigma = 1.5$, the three mentioned components are defined as [52]:

$$l(\mathbf{x}, \mathbf{y}) = \frac{2\mu_x\mu_y + C_1}{\mu_x^2 + \mu_y^2 + C_1} \tag{29}$$

$$c(\mathbf{x},\mathbf{y}) = \frac{2\sigma_x\sigma_y + C_2}{\sigma_x^2 + \sigma_y^2 + C_2} \quad (30)$$

$$s(\mathbf{x},\mathbf{y}) = \frac{\sigma_{xy} + C_3}{\sigma_x\sigma_y + C_3} \quad (31)$$

where $\mu_x$ and $\mu_y$ denote the spatial means within each local window, $\sigma_x^2$ and $\sigma_y^2$ represent the variance within each local window, and $\sigma_{xy}$ is the cross-covariance of windows. The constants $C_1 = 0.0001$, $C_2 = 0.0009$, and $C_3 = C_2/2$ prevent division by zero in each fraction.

For the introduced local windows $x$ and $y$, the SSIM is defined as [52]:

$$\text{SSIM}(\text{x},\text{y}) = l(\text{x},\text{y}) \cdot c(\text{x},\text{y}).s(\text{x},\text{y}) = \frac{(2\mu_x\mu_y + C_1)(2\sigma_{xy} + C_2)}{(\mu_x^2 + \mu_y^2 + C_1)(\sigma_x^2 + \sigma_y^2 + C_2)} \quad (32)$$

The final SSIM value is computed for a frame by averaging all the SSIM values obtained for the local windows across that frame.

MS-SSIM applies a similar computational process hierarchically at $M = 5$ different frame scales by employing a progressive downscaling process. MS-SSIM is computed as [52]:

$$\text{MS-SSIM}(\mathbf{x},\mathbf{y}) = [l_M(\mathbf{x},\mathbf{y})]^{\alpha_M} \cdot \prod_{m=1}^{M} [c_m(\mathbf{x},\mathbf{y}).s_m(\mathbf{x},\mathbf{y})]^{\beta_m} \quad (33)$$

where $\beta_1 = 0.0448$, $\beta_2 = 0.2856$, $\beta_3 = 0.3001$, $\beta_4 = 0.2363$, and $\alpha_M = \beta_M = 0.1333$.

Because MS-SSIM yields a bounded score between 0 and 1, where a higher score indicates better frame quality, our MS-SSIM loss component ($L_{\text{MS_SSIM}}$) is defined as:

$$L_{\text{MS_SSIM}} = 1 - \text{MS_SSIM}(\hat{I}_{t+1}, I_{t+1}) \quad (34)$$

Our total loss function, as a linear combination of the previously mentioned components, is defined as:

$$L_{\text{total}} = \alpha L_1 + \alpha L_{\text{GDL}} + (2 - 2\alpha) L_{\text{MS_SSIM}} \quad (35)$$

The coefficient $\alpha \in (0, 1)$ ensures all loss weights remain positive. Because $L_1$ and $L_{\text{GDL}}$ both penalize pixel-level errors by focusing on pixel intensities and spatial gradients, respectively, they share the same weight. This design explicitly partitions the objective function, weighing the combined pixel-level errors against the holistic structural integrity measured by $L_{\text{MS_SSIM}}$. The optimization process for the coefficient $\alpha$ is discussed in detail in Section 4.3.

### 3.6. Anomaly score

Our predictive model is trained exclusively on normal samples. This should result in high prediction errors for anomalous samples during model inference. To detect anomalies based on prediction errors, we define an anomaly scoring mechanism based on both Peak Signal-to-Noise Ratio (PSNR) and MS-SSIM. We integrate both metrics because PSNR effectively captures pixel-level prediction errors, whereas MS-SSIM quantifies structural deviations.

While our training objective utilizes L1 loss to preserve pixel-wise accuracy, we employ PSNR for anomaly scoring because its reliance on the squared L2 error quadratically magnifies large pixel-wise deviations, thereby making anomalous events more distinct. Given a predicted frame $\hat{I}_{t+1}$ and its corresponding ground truth $I_{t+1}$, PSNR is defined as:

$$\text{PSNR}\left(I_{t+1}, \hat{I}_{t+1}\right) = 10 \log_{10}\left(\frac{{I_{\max}}^2}{\left\|\hat{I}_{t+1} - I_{t+1}\right\|_2 / N}\right) \quad (36)$$

where $\|\,.\,\|$ denotes the L2 norm, $N$ is the total number of pixels in the frame, and $I_{\max}$ is the maximum possible pixel intensity value.

To measure structural frame quality across various scales, we compute the MS-SSIM score between $\hat{I}_{t+1}$ and $I_{t+1}$ as defined in Eq. (33). Consequently, anomaly scores $S_{\mathrm{PSNR}}(t)$ and $S_{\mathrm{MS_SSIM}}(t)$ are defined as:

$$S_{\mathrm{PSNR}}(t) = -\mathrm{PSNR}\left(I_{t+1}, \hat{I}_{t+1}\right) \tag{37}$$

$$S_{\mathrm{MS_SSIM}}(t) = 1 - \mathrm{MS_SSIM}\left(\hat{I}_{t+1}, I_{t+1}\right) \tag{38}$$

Subsequently, we apply Min-Max normalization independently to both $S_{\mathrm{PSNR}}(t)$ and $S_{\mathrm{MS_SSIM}}(t)$ within each individual video clip. Each of the obtained normalized scores, denoted as $\bar{S}_{\mathrm{PSNR}}(t)$ and $\bar{S}_{\mathrm{MS_SSIM}}(t)$, falls into the range $[0, 1]$:

$$\bar{S}_{\mathrm{PSNR}}(t) = \frac{S_{\mathrm{PSNR}}(t) - \min_{t'}(S_{\mathrm{PSNR}}(t'))}{\max_{t'}(S_{\mathrm{PSNR}}(t')) - \min_{t'}(S_{\mathrm{PSNR}}(t'))} \tag{39}$$

$$\bar{S}_{\mathrm{MS_SSIM}}(t) = \frac{S_{\mathrm{MS_SSIM}}(t) - \min_{t'}(S_{\mathrm{MS_SSIM}}(t'))}{\max_{t'}(S_{\mathrm{MS_SSIM}}(t')) - \min_{t'}(S_{\mathrm{MS_SSIM}}(t'))} \tag{40}$$

where $t'$ ranges over all frame indices within a given video clip.

Next, $\bar{S}_{\mathrm{PSNR}}(t)$ and $\bar{S}_{\mathrm{MS_SSIM}}(t)$ are combined linearly using the coefficient $w$:

$$S(t) = w\bar{S}_{\mathrm{PSNR}}(t) + (1 - w)\bar{S}_{\mathrm{MS_SSIM}}(t) \tag{41}$$

Lastly, to obtain the final anomaly score, we utilize a 1D Gaussian filter to smooth $S(t)$. The coefficient $w \in [0, 1]$ guarantees a valid convex combination of the normalized scores. The optimal value of this coefficient is analyzed further in Section 4.

# 4. Experiments

## 4.1. Datasets

To evaluate the performance of the MS$^2$CAST framework following a semi-supervised approach, we conduct experiments on three widely used, publicly available benchmark datasets: UCSD Ped2 [53], CUHK Avenue [54], and ShanghaiTech [55].

The UCSD Ped2 [53] dataset captures a pedestrian walkway in a single scene, where the training data consists of normal pedestrian walking. The test set contains 12 anomalous events, with anomalies defined as non-pedestrian objects, including bicycles, skateboards, and vehicles entering the walkway. The dataset includes 16 training videos and 12 testing videos with a resolution of $240 \times 360$ pixels.

The CUHK Avenue [54] dataset contains 16 training videos and 21 testing videos at a resolution of $360 \times 640$ pixels, recorded from a single scene of a campus avenue. The dataset includes 47 anomalous events such as individuals running, walking in abnormal directions, and throwing objects.

The ShanghaiTech [55] dataset is one of the most comprehensive and challenging benchmark datasets that can be used to evaluate multi-scene learning capability of a VAD model. It encompasses 13 distinct scenes, consisting of 330 training videos and 107 testing videos with a resolution of $480 \times 856$ pixels. The test set includes 130 abnormal events, covering a wide range of anomalies like people running, fighting, and riding. A comparative summary of the benchmark datasets is provided in Table 1.

**Table 1**
Comparative summary of the three benchmark datasets

| Dataset | #Videos | | #Frames | | #Anomalous Events | #Scenes |
|---|---|---|---|---|---|---|
| | Training | Testing | Training | Testing | | |
| UCSD Ped 2 | 16 | 12 | 2,550 | 2,010 | 12 | 1 |
| CUHK Avenue | 16 | 21 | 15,328 | 15,324 | 47 | 1 |
| ShanghaiTech | 330 | 107 | 274,515 | 42,883 | 130 | 13 |

## 4.2. Evaluation criterion and implementation details

Following established protocols in VAD research [9], we use the frame-level Receiver Operating Characteristic (ROC) Area Under the Curve (AUC) to quantitatively evaluate our proposed framework.

The MS$^2$CAST framework is implemented in PyTorch and evaluated on a single NVIDIA A100 GPU (40 GB VRAM). Each input video clip consists of four consecutive frames used to predict one future frame. Input frames are extracted and resized to a resolution of $224 \times 224$ pixels in RGB format, and their pixel values are scaled to a continuous range of $[0, 1]$. For training our model on each dataset, we use an AdamW optimizer with a weight decay of 0.01 and linear warmup followed by cosine annealing to schedule the learning rate. We use a batch size of 16 with a peak learning rate of $2 \times 10^{-4}$ over 25 epochs for the UCSD Ped2 dataset; a batch size of 32 with a peak learning rate of $5 \times 10^{-4}$ over 15 epochs for the CUHK Avenue dataset; and a batch size of 128 with a peak learning rate of $1 \times 10^{-3}$ over 5 epochs for the ShanghaiTech dataset.

## 4.3. Loss function optimization

As defined in Eq. (35), the hyperparameter $\alpha$ plays an integral role in determining the impact of each component within the proposed total loss function, which is a linear combination of three different criteria. To specify the optimal value for $\alpha$, we conducted experiments tuning $\alpha$ across the set $\{0.2, 0.3, 0.4, 0.5\}$ using the architecture featuring Post-LN Cross-Attention and the SwinLSTM temporal module. We evaluated the model on the CUHK Avenue and UCSD Ped2 datasets using $w = 1.0$ and $w = 0.6$, respectively, which are the optimal values for the coefficient $w$ in Eq. (41) for each dataset. Table 2 summarizes the frame-level AUC scores obtained for each value of $\alpha$.

The results highlight the importance of the perceptual loss component ($L_{\text{MS_SSIM}}$) over the pixel-wise loss component ($L_1$) and the GDL ($L_{\text{GDL}}$). Across both benchmark datasets, the architecture achieved its peak performance at $\alpha = 0.2$. Consequently, $\alpha = 0.2$ was selected as the hyperparameter for all subsequent experiments.

**Table 2**
Frame-level AUC (%) across different $\alpha$ values for the architecture with Post-Layer Normalization (Post-LN) cross-attention and the SwinLSTM temporal module on the UCSD Ped2 and CUHK Avenue datasets.

| $\alpha$ | CUHK Avenue ($w = 1.0$) | UCSD Ped2 ($w = 0.6$) |
|---|---|---|
| 0.2 | 86.4 | 96.2 |
| 0.3 | 85.6 | 95.8 |
| 0.4 | 85.4 | 93.5 |
| 0.5 | 85.1 | 95.6 |

## 4.4. Architectural ablations and performance evaluation

### 4.4.1. Spatial encoder and spatiotemporal module effect evaluation

To evaluate the effectiveness of our proposed multi-scale spatial cross-attention mechanisms and to compare the interchangeable spatiotemporal modules, we conducted a comprehensive ablation study. We compared the anomaly detection capabilities of ten different models on the UCSD Ped2 and CUHK Avenue datasets. For each variant of the spatiotemporal module, we tested five distinct spatial encoders: a multi-scale Pre-LN cross-attention encoder, a multi-scale Post-LN cross-attention encoder, a pre-trained MViTv2-S encoder, a fine-tuned MViTv2-S encoder, and a pre-trained ConvNeXt-T encoder.

In addition to comparing the pre-trained MViTv2-S as a single spatial encoder against the encoders that are using the two variants of the spatial cross-attention mechanism, we investigated the effect of fine-tuning the pre-trained MViTv2-S encoder for domain adaptation. We adopted a layer-wise learning rate strategy for this fine-tuning process. To stabilize the spatiotemporal module and the decoder, the MViTv2-S backbone was kept frozen during the initial training warmup steps. Following this warmup, the pre-trained model was unfrozen to begin fine-tuning, utilizing a block-wise learning rate decay factor of 0.9. Under this strategy, the deepest block of the pre-trained model received the lowest learning rate, while the shallowest block shared the same learning rate as the spatiotemporal module and the decoder. In all evaluated single-encoder architectures, the three integrated skip connections originate from the intermediate stages of the respective encoder. Table 3 and Table 4 present the frame-level AUC comparisons for each evaluated architecture across the UCSD Ped2 and CUHK Avenue datasets, respectively.

**Table 3**
Frame-level AUC (%) comparison across different combinations of spatial encoders and spatiotemporal modules on UCSD Ped2 dataset, where $w$ is the coefficient used to calculate the anomaly score.

| Spatiotemporal Module | Spatial Encoder | Optimal $w$ | AUC (%) |
|---|---|---|---|
| VMRNN | Pre-trained ConvNeXt-T | 0.5 | 95.2 |
| | Pre-trained MViTv2-S | 0.5 | 95.5 |
| | Fine-tuned MViTv2-S | 0.5 | 95.1 |
| | Pre-LN cross-attention | 0.6 | 96.5 |
| | Post-LN cross-attention | 0.5 | 95.7 |
| SwinLSTM | Pre-trained ConvNeXt-T | 0.3 | 95.7 |
| | Pre-trained MViTv2-S | 0.6 | 96.0 |
| | Fine-tuned MViTv2-S | 0.6 | 95.6 |
| | Pre-LN cross-attention | 0.6 | 96.1 |
| | Post-LN cross-attention | 0.6 | 96.2 |

**Table 4**
Frame-level AUC (%) comparison across different combinations of spatial encoders and spatiotemporal modules on CUHK Avenue dataset, where $w$ is the coefficient used to calculate the anomaly score.

| Spatiotemporal Module | Spatial Encoder | Optimal $w$ | AUC (%) |
|---|---|---|---|
| VMRNN | Pre-trained ConvNeXt-T | 0.9 | 84.6 |
| | Pre-trained MViTv2-S | 0.8 | 86.2 |
| | Fine-tuned MViTv2-S | 0.8 | 86.1 |
| | Pre-LN cross-attention | 0.8 | 85.6 |
| | Post-LN cross-attention | 0.9 | 85.6 |
| SwinLSTM | Pre-trained ConvNeXt-T | 0.8 | 85.6 |
| | Pre-trained MViTv2-S | 0.8 | 86.1 |
| | Fine-tuned MViTv2-S | 0.7 | 84.8 |
| | Pre-LN cross-attention | 0.9 | 85.8 |
| | Post-LN cross-attention | 1.0 | 86.4 |

In all cases, fine-tuning the MViTv2-S model pre-trained on the ImageNet-1k [48] dataset resulted in AUC degradation. This decline is likely because fine-tuning a high-capacity pre-trained model requires a substantial amount of data, and fine-tuning such models with an insufficient number of samples can potentially disrupt the generalized feature representations initially learned from the massive ImageNet-1k dataset. As reported in Table 1, the CUHK Avenue and UCSD Ped2 datasets have much fewer training frames than the number of images in the ImageNet-1k dataset, which contains more than one million samples across 1,000 classes in its training set [48].

The results presented in this section exhibit different trends across these datasets. On the UCSD Ped2 dataset, using cross-attention mechanisms for the spatial encoder consistently resulted in a higher AUC score. Conversely, on the CUHK Avenue dataset, utilizing VMRNN as the spatiotemporal module and incorporating cross-attention mechanisms in the spatial encoder to leverage both MViTv2-S and ConvNeXt-T did not improve the AUC over using MViTv2-S alone. Notably, the configuration pairing the multi-scale Post-LN cross-attention spatial encoder with a SwinLSTM as the spatiotemporal module achieved the highest AUC on the CUHK Avenue dataset across all ten evaluated configurations, whereas using the multi-scale Pre-LN spatial cross-attention encoder demonstrated inferior performance compared with employing pre-trained MViTv2-S alone.

Although the model that incorporates the multi-scale Post-LN spatial cross-attention encoder and SwinLSTM exhibited superior performance across all examined models on both datasets, we further evaluated all configurations in noisy environments to assess their noise resilience and robustness. The noise resilience of these models is discussed in Section 4.4.2.

#### 4.4.2. Robustness to noise

Surveillance systems commonly suffer from environmental noise that adversely affects the quality of their recordings [56]. Consequently, it is crucial to consider the noise resilience of a VAD model as a factor in model selection for such applications. To assess the performance of the models introduced in Section 4.4.1 under various noisy conditions, we added zero-mean additive white Gaussian noise (AWGN) to the input frames across different standard deviations ($\sigma$) during model evaluation. All other model evaluation protocols remained identical to those described in Section 4.4.1. Table 5 and Table 6 present the frame-level AUC of each model under various AWGN standard deviations ($\sigma$) on the UCSD Ped2 and CUHK Avenue datasets, respectively.

**Table 5**
Frame-level AUC (%) comparison under various additive white Gaussian noise (AWGN) standard deviations ($\sigma$) on the UCSD Ped2 dataset.

| Spatiotemporal Module | Spatial Encoder | AUC (%) | | | |
|---|---|---|---|---|---|
| | | No Noise | $\sigma = 5$ | $\sigma = 10$ | $\sigma = 15$ |
| VMRNN | Pre-trained ConvNeXt-T | 95.2 | 94.9 | 89.4 | 69.8 |
| | Pre-trained MViTv2-S | 95.5 | 93.2 | 91.2 | 82.7 |
| | Pre-LN cross-attention | 96.5 | 95.4 | 92.8 | 89.4 |
| | Post-LN cross-attention | 95.7 | 95.1 | 93.5 | 85.0 |
| SwinLSTM | Pre-trained ConvNeXt-T | 95.7 | 94.5 | 85.0 | 72.6 |
| | Pre-trained MViTv2-S | 96.0 | 95.5 | 90.7 | 87.6 |
| | Pre-LN cross-attention | 96.1 | 95.9 | 93.6 | 90.5 |
| | Post-LN cross-attention | 96.2 | 95.5 | 95.2 | 91.9 |

**Table 6**
Frame-level AUC (%) comparison under various additive white Gaussian noise (AWGN) standard deviations ($\sigma$) on the CUHK Avenue dataset.

| Spatiotemporal Module | Spatial Encoder | AUC (%) | | | |
|---|---|---|---|---|---|
| | | No Noise | $\sigma = 5$ | $\sigma = 10$ | $\sigma = 15$ |
| VMRNN | Pre-trained ConvNeXt-T | 84.6 | 84.3 | 82.5 | 81.5 |
| | Pre-trained MViTv2-S | 86.2 | 86.2 | 85.9 | 82.9 |
| | Pre-LN cross-attention | 85.6 | 85.6 | 83.2 | 79.9 |
| | Post-LN cross-attention | 85.6 | 84.1 | 79.2 | 75.1 |
| SwinLSTM | Pre-trained ConvNeXt-T | 85.6 | 85.1 | 82.2 | 78.2 |
| | Pre-trained MViTv2-S | 86.1 | 86.1 | 84.5 | 81.9 |
| | Pre-LN cross-attention | 85.8 | 85.8 | 85.4 | 82.4 |
| | Post-LN cross-attention | 86.4 | 84.2 | 77.5 | 68.7 |

Across all configurations, the AUC monotonically decreases as the AWGN standard deviation ($\sigma$) increases. However, the severity of AUC degradation varies depending on the configuration.

Although pairing the multi-scale Post-LN cross-attention encoder with SwinLSTM outperforms the other configurations in environments without noise, the AUC of this configuration on the CUHK Avenue dataset drops significantly under noisy conditions. Conversely, using the multi-scale Pre-LN cross-attention encoder along with SwinLSTM generally exhibits highly robust anomaly detection performance in the presence of noise on both datasets. This difference in noise resilience stems from how layer normalization interacts with feature representations within the integrated cross-attention mechanism. In the Post-LN cross-attention mechanism (Eqs. (7)–(10)), layer normalization operates on the residual sum obtained after the scaled dot-product attention (Eq. (5)), which is computed from unnormalized, noisy Query and Key/Value representations. Consequently, this can lead to a distorted attention weight matrix that is summed with the noisy representation of the Query. In contrast, the Pre-LN cross-attention mechanism (Eqs. (11)–(14)) normalizes the Query and Key/Value representations prior to the computation of the attention weight matrix and the subsequent residual summation. This pre-normalization statistically standardizes feature maps before the cross-attention computations, resulting in feature representations that are less disrupted by noise.

It is worth noting that pairing the pre-trained MViTv2-S encoder with VMRNN shows robustness on the CUHK Avenue dataset, even though the configuration that uses the multi-scale Pre-LN cross-attention encoder along with SwinLSTM consistently outperforms it on the UCSD Ped2 dataset. Overall, our findings confirm that pairing the multi-scale Pre-LN cross-attention encoder with SwinLSTM results in a robust architecture capable of maintaining high AUC scores under various noisy conditions across both datasets. Consequently, we selected this specific configuration to evaluate its multi-scene learning capability on the ShanghaiTech dataset, which is a comprehensive, multi-scene benchmark.

#### 4.4.3. Computational efficiency analysis

Beyond anomaly detection capability and noise resilience, operational efficiency is a critical factor for deploying VAD models in real-world scenarios. To assess the computational efficiency of our proposed architecture during inference and compare it with other alternatives introduced in our ablation studies, we measured latency (milliseconds), throughput (frames per second), and peak GPU memory usage (MB) for six distinct architectural configurations. Table 7 compares the computational efficiency of the evaluated architectures using three different metrics.

**Table 7**
Computational efficiency comparison across different combinations of spatial encoders and spatiotemporal modules.

| Spatiotemporal Module | Spatial Encoder | Latency (ms) | Throughput (FPS) | Peak GPU Memory (MB) |
|---|---|---|---|---|
| VMRNN | Pre-trained ConvNeXt-T | 38 | 161 | 536 |
| | Pre-trained MViTv2-S | 59 | 106 | 617 |
| | Pre-LN cross-attention | 72 | 64 | 750 |
| SwinLSTM | Pre-trained ConvNeXt-T | 31 | 192 | 592 |
| | Pre-trained MViTv2-S | 51 | 118 | 674 |
| | Pre-LN cross-attention | 62 | 68 | 806 |

The results demonstrate a clear trade-off between anomaly detection capability and computational efficiency across different architectural configurations. Using a single-encoder configuration yields lower latency and higher throughput compared with the configuration that employs a multi-scale cross-attention encoder. However, as established in Section 4.4.1 and Section 4.4.2, relying on a single-encoder configuration limits the anomaly detection capability of a model. A comparison of the two interchangeable spatiotemporal modules reveals that SwinLSTM consistently operates with lower latency and higher throughput than VMRNN across all configurations, though it requires more GPU memory.

The peak GPU memory usage of our selected architecture that uses the multi-scale Pre-LN cross-attention encoder along with SwinLSTM is 806 MB. This indicates that our framework can successfully maintain peak GPU memory usage under 1 GB while demonstrating strong anomaly detection capability.

## 4.5. Comparison with existing methods

To further evaluate the multi-scene learning capability of our selected configuration for the MS$^2$CAST framework, which employs the multi-scale Pre-LN cross-attention encoder along with SwinLSTM, we conducted experiments on the ShanghaiTech dataset. On this benchmark, the configuration achieved an AUC of 75.3% using an anomaly score coefficient $w = 0.2$. Table 8 compares the frame-level AUC scores of MS$^2$CAST against top-tier semi-supervised VAD methods across the three benchmark datasets, categorizing these methods into multi-modal and single-modal techniques. Our proposed framework achieves performance comparable to these methods on the UCSD Ped2 and CUHK Avenue datasets, while outperforming all of them on the challenging ShanghaiTech dataset.

**Table 8**
Frame-level AUC (%) comparison with top-tier semi-supervised video anomaly detection methods on the UCSD Ped2, CUHK Avenue, and ShanghaiTech datasets.

| Type | Method | UCSD Ped2 | CUHK Avenue | ShanghaiTech |
|---|---|---|---|---|
| Multi-Modal | DeepOC [30] | 96.9 | 86.6 | _ |
| | FSCN [32] | 92.8 | 85.5 | _ |
| | AMMC-Net [34] | 96.6 | 86.6 | 73.7 |
| | MAMC [33] | 96.7 | **87.6** | 71.5 |
| | CF²M-Net [16] | **96.9** | 85.7 | 73.9 |
| Single-Modal | ConvLSTM-AE [40] | 88.1 | 77.0 | _ |
| | TransAnomaly [21] | 96.4 | 87.0 | _ |
| | MNAD-R [5] | 90.2 | 82.8 | 69.8 |
| | MemAE [4] | 94.1 | 83.3 | 71.2 |
| | Frame-Pred [9] | 95.4 | 85.1 | 72.8 |
| | Zhang et al. [6] | 95.8 | 84.9 | 71.4 |
| | F²PN [11] | 96.2 | 85.7 | 73.0 |
| | MGAN-CL [14] | 96.5 | 87.1 | 73.6 |
| | Dual GroupGAN [13] | 96.6 | 85.5 | 73.1 |
| | Multispace [12] | 95.4 | 86.8 | 73.6 |
| | **MS²CAST** (Pre-LN + SwinLSTM) | 96.1 | 85.8 | **75.3** |

Although MAMC [33] and CF$^2$M-Net [16] demonstrate strong performance on the single-scene datasets, they utilize pre-trained optical flow networks that introduce computational overhead. Additionally, they both incorporate memory modules, which can lead to poor cross-scene generalization [6]. Among single-modal techniques, Dual GroupGAN [13] and MGAN-CL [14] achieve high AUC scores on the single-scene datasets. However, their reliance on adversarial training can pose challenges regarding training stability and hyperparameter tuning [15]. In contrast, our end-to-end framework demonstrates superior multi-scene learning capability on the ShanghaiTech dataset and exhibits competitive anomaly detection performance across the single-scene benchmarks. Moreover, MS$^2$CAST maintains peak GPU memory usage under 1 GB and avoids the computational overhead associated with adversarial training strategies and heavy pre-trained optical flow models.

To qualitatively demonstrate the effectiveness of the MS$^2$CAST framework, Figs. 4(a), 4(b), and 4(c) visualize the actual frame, the predicted frame, and the corresponding error map during anomalous events across the UCSD Ped2, CUHK Avenue, and ShanghaiTech datasets, respectively. Each error map shows the pixel-wise deviations between the prediction and actual frame. Furthermore, Figs. 5(a), 5(b), and 5(c) illustrate the anomaly score over time across specific video clips for these respective datasets, while the red background indicates ground truth anomalous frames.

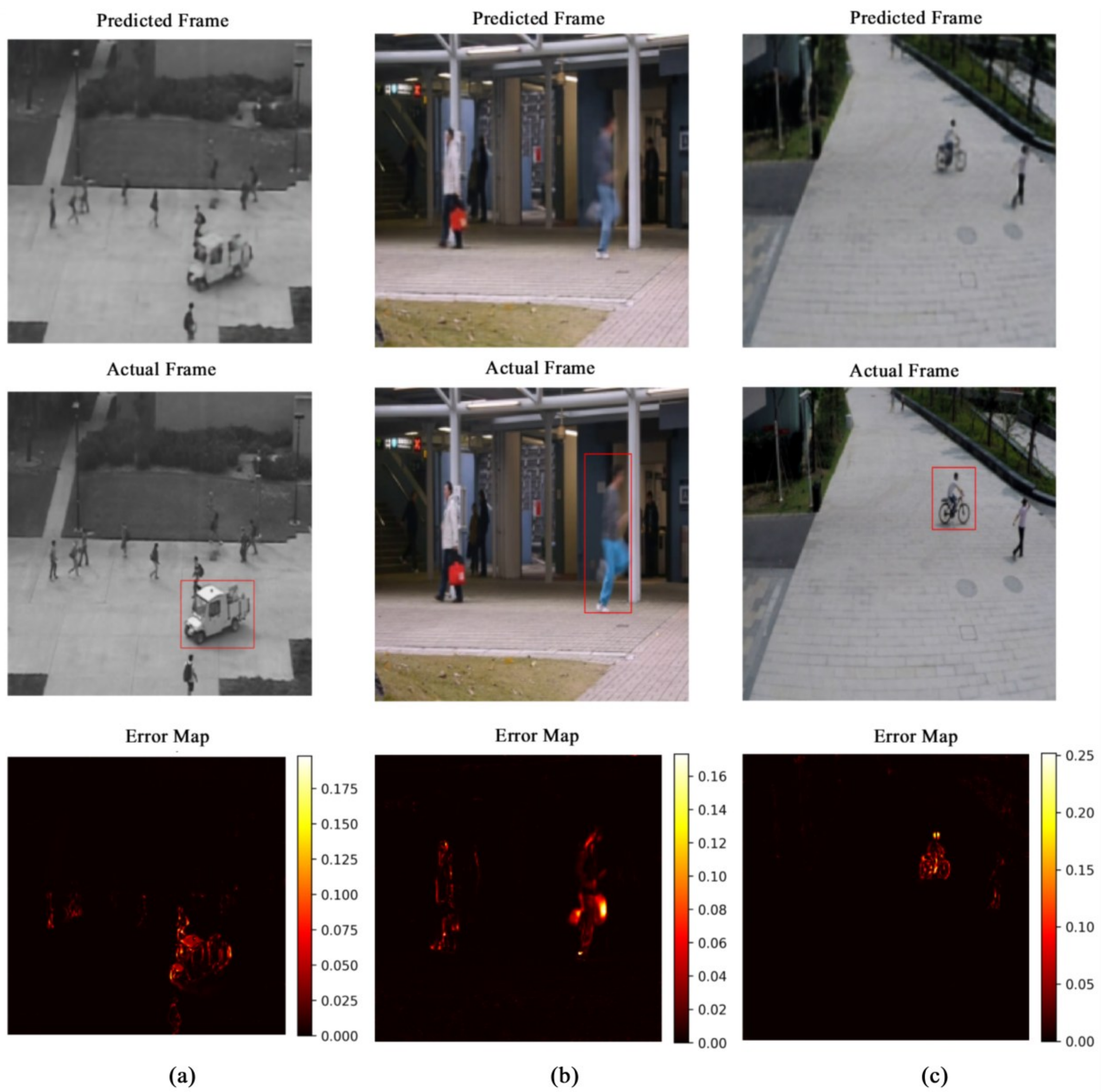

**Fig. 4.** Visualization of the actual frame, the corresponding predicted frame, and the pixel-wise error map during anomalous events across the (a) UCSD Ped2, (b) CUHK Avenue, (c) ShanghaiTech datasets.

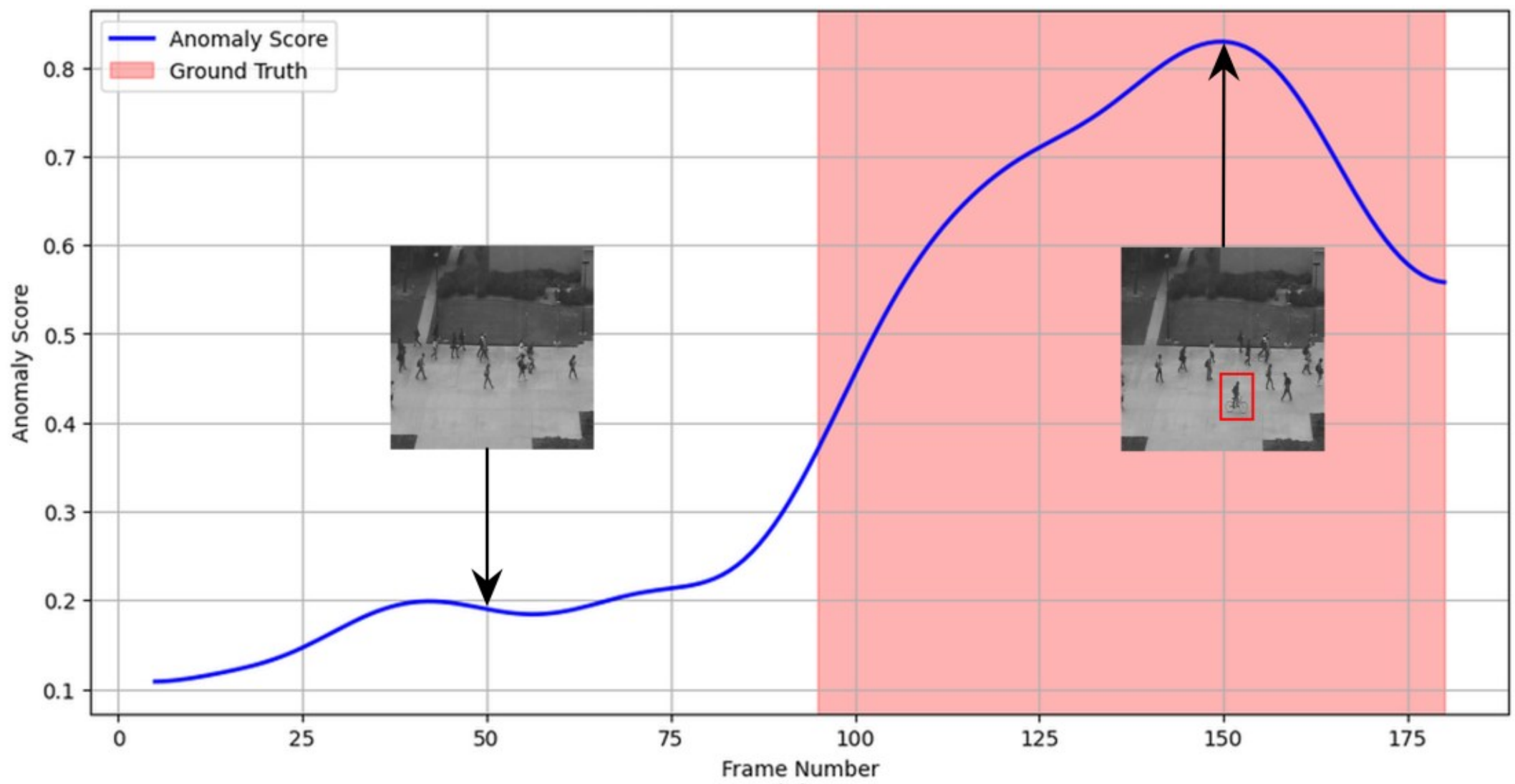

(a)

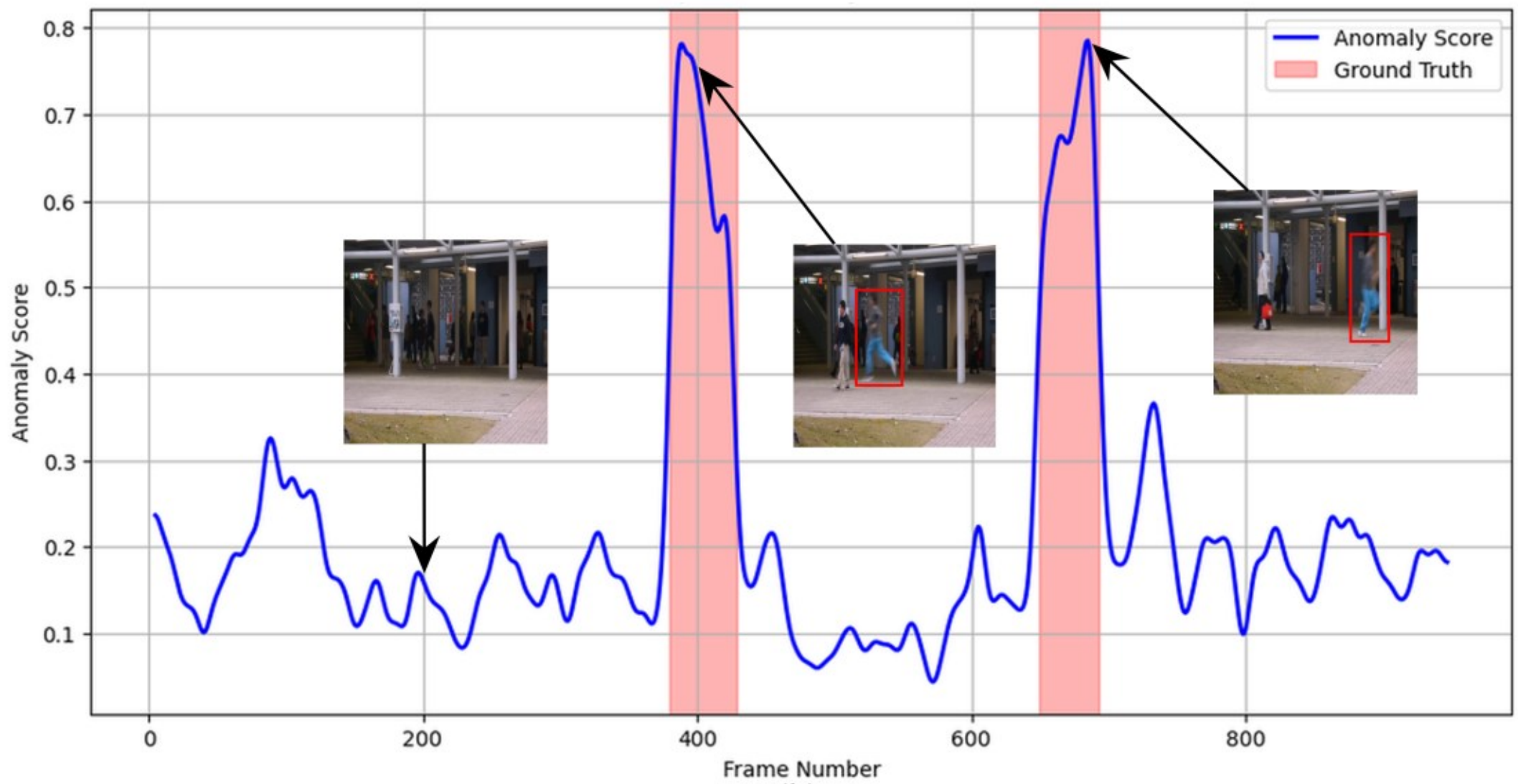

(b)

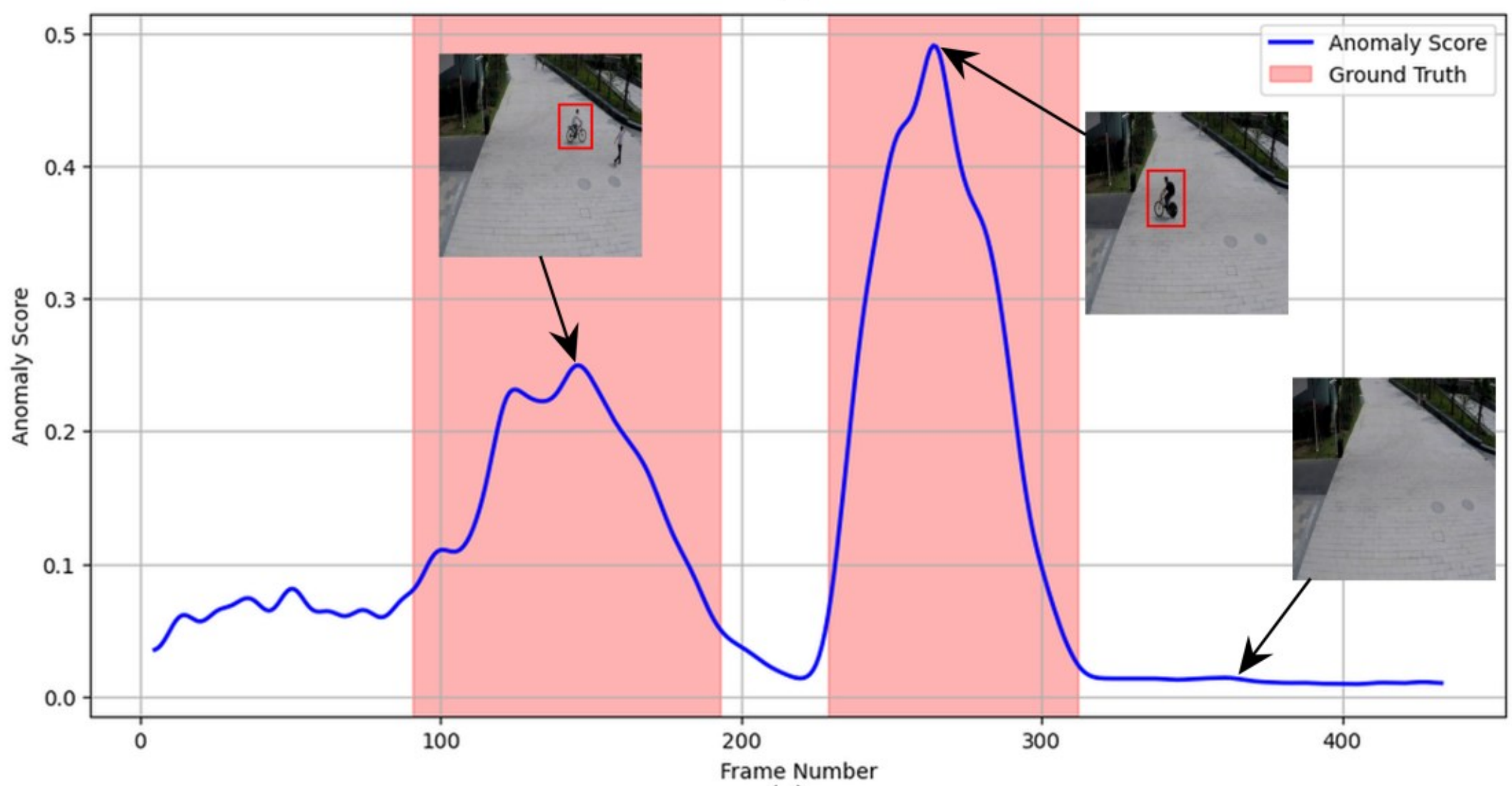

(c)

**Fig. 5.** Anomaly scores plotted over time for sample video clips from the (a) UCSD Ped2, (b) CUHK Avenue, and (c) ShanghaiTech datasets. Ground truth anomalous frames are highlighted by the red background.

### 4.6. Hyperparameter sensitivity analysis

As defined in Eq. (41), the final anomaly score is computed as a weighted linear combination of the normalized PSNR-based score and the normalized MS-SSIM-based score. The relative contribution of each component depends on the hyperparameter $w$, where the final anomaly score with $w = 0$ exclusively relies on MS-SSIM, and the score with $w = 1.0$ relies solely on PSNR. To investigate the effect of this hyperparameter on the anomaly detection capability of the selected configuration for MS$^2$CAST, which employs the multi-scale Pre-LN cross-attention encoder along with SwinLSTM, we conducted a sensitivity analysis by varying $w$ from 0 to 1 in increments of 0.1. The frame-level AUC scores for each value of the coefficient $w$ are illustrated in Fig. 6 for the UCSD Ped2 (Fig. 6(a)), CUHK Avenue (Fig. 6(b)), and ShanghaiTech (Fig. 6(c)) datasets. The results demonstrate that the relationship between the frame-level AUC and $w$ is highly dependent on the dataset, yielding different optimal values of $w$ for each benchmark.

On the UCSD Ped2 dataset, as $w$ increases from 0, the AUC monotonically increases from 93.9%, achieving its highest value of 96.1% at $w = 0.6$. The AUC score remains unchanged at $w = 0.7$, $w = 0.8$, and $w = 0.9$ before slightly decreasing to 96.0% at $w = 1.0$. Similarly, on the CUHK Avenue dataset, the AUC exhibits a consistent upward trend, starting from 79.7% at $w = 0$ and reaching a maximum of 85.8% at $w = 0.9$, followed by a slight decrease to 85.7% at $w = 1.0$. Unlike the single-scene datasets, the AUC score on the ShanghaiTech dataset initially increases, reaching its maximum value at $w = 0.2$ and sustaining this score at $w = 0.3$ and $w = 0.4$. After this point, further increments of $w$ yield a monotonic decrease in the AUC score, ultimately reaching a value of 73.9% at $w = 1.0$.

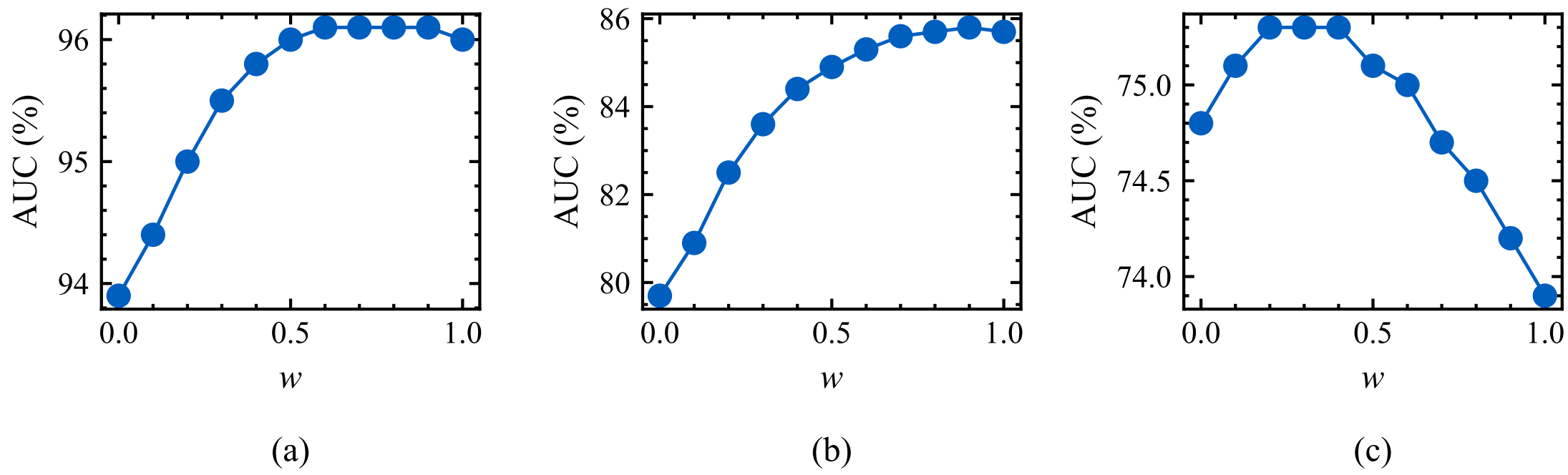

**Fig. 6.** Frame-level AUC (%) of the selected configuration for MS$^2$CAST across different values of the coefficient $w$, evaluated on the (a) UCSD Ped2, (b) CUHK Avenue, and (c) ShanghaiTech datasets.

Generally, assigning a higher weight to MS-SSIM to determine the final anomaly score yields higher AUC scores on the ShanghaiTech dataset, whereas assigning a higher weight to PSNR results in the maximum AUC scores on the CUHK Avenue and UCSD Ped2 datasets. These dataset-dependent variations confirm the necessity of our dual-metric anomaly scoring mechanism, which provides the flexibility required to adapt to the specific spatial and temporal characteristics of different surveillance environments.

## 5. Conclusion

In this paper, we introduce $MS^2CAST$ an end-to-end semi-supervised predictive VAD framework. This framework fuses the distinct hierarchical feature representations of a ViT and a CNN via a multi-scale cross-attention mechanism to encode spatial features. This spatial encoding strategy is paired with an efficient spatiotemporal module to capture motion dynamics without relying on computationally demanding deep optical flow networks. Furthermore, the objective function for this framework focuses on both pixel-wise accuracy and structural integrity while enforcing sharp edges in the predicted frames through spatial gradients. Our comprehensive ablation studies revealed several critical insights regarding architectural design choices, confirming the effectiveness of leveraging the multi-scale spatial cross-attention mechanism in the spatial encoder. We demonstrated that employing the multi-scale Pre-LN cross-attention encoder along with a SwinLSTM spatiotemporal module yields competitive anomaly detection performance on the UCSD Ped2 and CUHK Avenue datasets while outperforming top-tier VAD methods on the ShanghaiTech dataset. This architectural configuration maintains peak GPU memory usage under 1 GB while exhibiting strong robustness across various types of additive noise. While the anomaly scoring currently exhibits dataset-dependent sensitivity to the hyperparameter $w$, future work will focus on exploring alternative metrics and formulating a unified, parameter-free anomaly score function.